\documentclass{article}

\usepackage{microtype}
\usepackage{graphicx}
\usepackage{subcaption}
\usepackage{booktabs} % for professional tables
\usepackage{bm}
\usepackage[strings]{underscore}
\usepackage[dvipsnames]{xcolor}
\usepackage{amssymb}

\usepackage{hyperref}

\let\savedaddcontentsline\addcontentsline
\usepackage[preprint]{icml2026}

\usepackage{amsmath}
\usepackage{amssymb}
\usepackage{mathtools}
\usepackage{amsthm}

\usepackage[capitalize,noabbrev]{cleveref}

\theoremstyle{plain}
\newtheorem{theorem}{Theorem}

\newtheorem{lemma}{Lemma}
\newtheorem{corollary}{Corollary}

\newtheorem{assumption}{Assumption}

\icmltitlerunning{Compressed Simulation-Based Inference}

\begin{document}

\twocolumn[
  \icmltitle{Compressed Simulation-Based Inference: Neural Posterior\\ Estimation for Massive Data via Low-Dimensional Sketches}  
%%% Note: previous title was "Scalable Neural Posterior Estimation via Low-Dimensional Sketches"


  \icmlsetsymbol{equal}{*}

  \begin{icmlauthorlist}
    \icmlauthor{Michael Wieck-Sosa}{yyy}
    \icmlauthor{Cosma Rohilla Shalizi}{yyy}
  \end{icmlauthorlist}

  \icmlaffiliation{yyy}{Department of Statistics \& Data Science, Carnegie Mellon University, Pittsburgh, PA, USA}

  \icmlcorrespondingauthor{Cosma Rohilla Shalizi}{cshalizi@cmu.edu}

  \icmlkeywords{Machine Learning, ICML}

  \vskip 0.3in
]



\printAffiliationsAndNotice{}  % no special notice (required even if empty)

\begin{abstract}
     Neural posterior estimation (NPE) is a popular Bayesian simulation-based inference (SBI) approach that uses neural networks trained on simulated data--parameter pairs to approximate the posterior given a user-chosen prior. For high-dimensional observations, NPE methods typically approximate the conditional density of the parameter given a user-chosen or learned summary statistic, rather than the entire dataset, due to the curse of dimensionality. However, both approaches face challenges: hand-crafting good summaries for complex models requires considerable effort from scientists, while learning them from massive datasets can be prohibitively computationally expensive. To address these challenges, we propose approximating the conditional density of a parameter given low-dimensional sketches of the data. We show that, for a model with a $p$-dimensional parameter, only $2p+1$ random sketches are needed for posterior consistency. We compare with state-of-the-art embedding networks for NPE, and we find that our method \emph{matches} the best embedding network's performance while \emph{increasing computational efficiency}.
\end{abstract}
% abstract 4-6 sentences
% main paper (excluding references and appendices) 8 pages (final version 9 pages)
% appendix as long as you want (total file size may not exceed 10MB)
% single pdf: main, references, appendix
\section{Introduction}\label{section:introduction}

%%% Note: I'm saying fixed here to emphasize that we're studying this problem from a frequentist perspective

We observe an $\mathbb{R}^d$-valued sequence $X_t^{\mathrm{obs}}$, $t=1,\ldots,n$. As frequentists, we assume that the data come from a model with an unknown \textit{fixed} parameter $\theta_0\in\Theta \subset \mathbb{R}^p$. We would like to infer the true parameter $\theta_0$ based on the observations.

Unfortunately, the likelihood of this model is intractable, which rules out conventional likelihood-based techniques. However, simulating length-$n$ sequences, $(X_{1:n}^{(r)}(\theta))_{r\in [B]}$, from this model at any value of $\theta\in\Theta$ is straightforward. For some summary statistic $S_n(\cdot)$, we aim to approximate the conditional density $\pi_n(\theta \mid S_n(X_{1:n}^{\mathrm{obs}}))$ under a user-chosen \textit{reference density} $\pi(\theta)$.
%%% Note: on the parameter space $\Theta$.

The core idea of this paper is that we should replace hand-crafted and learned summary statistics, $S_n(\cdot)$, with time-averages of a small number of \textit{random features}. Using more features generally provides more information about the parameters (indeed, a common goal in neural SBI is learning a sufficient statistic of the data). However, our goal is to match the performance of state-of-the-art embedding networks at a \textit{lower computational cost} by using low-dimensional sketches of the data.

We show that $2p+1$ random sketches of the data are sufficient for posterior consistency for a model with a $p$-dimensional parameter $\theta$. Inspired by these theoretical results, we propose a new NPE method based on sketching that is \textit{agnostic} to the sample size $n$ and observation dimension $d$, enabling our method to scale to massive datasets. Our experiments in Section~\ref{section:experiments} show that NPE methods can be made more scalable via \textit{randomized} summary statistics.

\subsection{Related Work}\label{subsection:related_work}

We provide an overview of existing approaches for summary statistics (``hand-crafted'' and ``learned'') in SBI, and we contrast them with our proposed approach (``randomized'').

\begin{table}[h!]
\centering
\caption{Comparison of paradigms for summary statistics. The table displays yes ($\checkmark$), depends on method ($\sim$), and no ($\times$).}
\label{tab:summary_statistics}
\footnotesize
\setlength{\tabcolsep}{4pt}
\begin{tabular}{@{}lccc@{}}
\toprule
& Scalable & Automated & Interpretable \\
\midrule
Hand-crafted & $\checkmark$ & $\times$ & $\checkmark$ \\
Learned      & $\sim$ & $\checkmark$ & $\times$ \\
Randomized   & $\checkmark$ & $\checkmark$ & $\times$ \\
\bottomrule
\end{tabular}
\end{table}

Neural SBI \citep{cranmer_sbi_review,zammitmangion_sbi_review} is a modern form of SBI that uses neural networks (typically) to approximate the posterior or likelihood. Traditionally, user-chosen summary statistics were a key ingredient of SBI \citep{mcfadden_method_sim_moments,indirect_inference_gourieroux_monfort_1993,sim_based_econometrics_gourieroux_monfort,wood_likelihood,BSL_price_2018,ABC_Beaumont_et_al_2002,ABC_Beaumont_2010,ABC_Beaumont_2019}.  However, \textit{hand-crafting} summary statistics requires expertise and can be challenging for complex models, often leading to ad-hoc choices \citep{Drovandi2022}.

More recently, approaches have been developed for \textit{learning} summary statistics either directly \citep{Chan2018,Wiqvist2019,Dyer2021,Pacchiardi2022} or jointly with the posterior in an end-to-end fashion \citep{Papamakarios2016,Greenberg2019,Hermans2020}. For more discussion, see \citet{Chen2023}. Both of these approaches are computationally demanding for simulators of massive datasets with large $n$ or $d$.

Motivated by these challenges, we propose a new approach to NPE based on \textit{randomized} summary statistics that provides an alternative to the two aforementioned paradigms. Random Fourier features \citep{Rahimi_Recht_2007_random_features,Rahimi_Recht_2008_weighted_sums,Rahimi_Recht_2008_uniform_approximation} have been widely used in statistics and machine learning. However, the way we use them is very different from how they are typically used, i.e., for approximating kernel methods. Our approach is largely inspired by recent results from \citet{random_features_estimation_paper,random_features_identification_paper,wiecksosa2026generativesequencemodelinginfinite}, although they do not consider the use of random Fourier features for neural SBI.

%\subsection{Our Contributions}

%The contributions of the paper are: 
%\begin{enumerate}
%    \item 
%\end{enumerate}

\section{Methods and Main Ideas}\label{section:methods_main_ideas}

\textsc{Notation}: For real numbers $x,y\in\mathbb{R}$, denote $x\lor y = \max(x,y)$.
For a vector $x\in\mathbb{R}^d$, denote the $\ell^q$ norm by $\left\|x\right\|_q$ and the Euclidean norm by $\left\|x\right\| = \left\|x\right\|_2$. For a random vector $X$ with distribution depending on $\theta\in\Theta$, $\left\|X\right\|_{\mathcal{L}^q(\theta)}=\left(\mathbb{E}_{\theta}\left\|X\right\|^q\right)^{1/q}$, where $\mathbb{E}_{\theta}(\cdot)$ denotes expectation with respect to this $\theta$-dependent distribution. For any natural number $j\in\mathbb{N}$, denote $[j]=\{1,2,\ldots,j\}$. For a sequence of random vectors $X_t$, $t=1,\ldots,n$, each taking values in $\mathbb{R}^d$, we denote a subsequence as $X_{t_1:t_2}=(X_{t_1},\ldots,X_{t_2})^{\top}$ for some $t_1,t_2 \in [n]$, $t_1\leq t_2$. For a sequence of random vectors $X_{1:n}(\theta)$, we denote its distribution by $P_{\theta,n}$. We may suppress the superscript or parameter when it is clear from the context.

%%% Note: \mathcal{L}^q(\theta) instead of L^q because using L for lag 

\subsection{Random Fourier Features}\label{subsection:random_Fourier_features}

We randomly draw $k$ functions $\varphi_1,\ldots,\varphi_k$ from a distribution over some function class $\mathcal{F}$, where each function $\varphi_i:\mathbb{R}^{(m+1) \times d}\xrightarrow[]{} \mathbb{R}$ accepts a length $m+1$ subsequence of the $d$-dimensional sequence as input. The lag order $m$ is chosen by the user based on the transition function of the model, e.g., set $m=m^{\ast}$ for an order $m^{\ast}$ Markov process.

Let $x_1,\ldots,x_{m+1}$ be vectors in $\mathbb{R}^d$, and let $x=(x_1,\ldots,x_{m+1})^{\top}\in\mathbb{R}^{(m+1)\times d}$. We consider functions from the class of random Fourier features, given by 
\begin{equation}\label{eqn:random_Fourier_features_order_m}\varphi_{i}(x)=\cos\left(\sum_{j=1}^{m+1} \Omega_{i,j} \cdot x_{j} + \alpha_i\right),\end{equation} where $\Omega_{i,j} \overset{\mathrm{iid}}{\sim} N(0,I_d)$, $\alpha_i \overset{\mathrm{iid}}{\sim} U(-\pi,\pi)$ for $i=1,\ldots,k$ and $j=1,\ldots,m+1$. Denote the parameters for $\varphi_i$ by \begin{equation}\label{eqn:parameters_for_the_random_features}\vartheta_i=(\Omega_{i,1}^{\top},\ldots,\Omega_{i,m+1}^{\top},\alpha_i)^{\top},\end{equation} and let $\vartheta=(\vartheta_1,\ldots,\vartheta_k)$. Denote all $k$ functions by \begin{equation}\label{eqn:all_k_random_Fourier_features_order_m}\varphi=(\varphi_1,\ldots,\varphi_k):\mathbb{R}^{(m+1) \times d}\xrightarrow[]{} \mathbb{R}^k.\end{equation}

\subsection{Algorithm}

Algorithm~\ref{algo:random_feature_neural_posterior_estimation} presents our method. We suggest that the number of features $k \geq 2p+1$ be chosen as a hyperparameter, analogous to an embedding dimension in an embedding network. In general, the user may decide how to choose the hyperparameters, e.g., by running Algorithm~\ref{algo:random_feature_neural_posterior_estimation} on held-out simulated summary--parameter pairs, and choosing the hyperparameters that maximize the average log-likelihood.

\begin{algorithm}[h!]
  \caption{NPE with Low-Dimensional Sketches}
  \label{algo:random_feature_neural_posterior_estimation}
  \begin{algorithmic}
    \STATE {\bfseries Input:} Observations $(X_t^{\mathrm{obs}})_{t=1}^n$, random Fourier features function class, reference density $\pi(\theta)$, function class $\mathcal{M}$
    \STATE Select hyperparameters of estimator 
    \STATE Obtain $S_n(X_{1:n}^{\mathrm{obs}})=\frac{1}{n-m}\sum_{t=m+1}^n \varphi(X_{t-m:t}^{\mathrm{obs}})$ 
    \FOR{$r=1$ {\bfseries to} $B$}
    \STATE Sample $\theta^{(r)}\sim \pi(\theta)$ and simulate $X_{1:n}^{(r)}\sim P_{\theta^{(r)},n}$
    \STATE Obtain $S_n(X_{1:n}^{(r)})=\frac{1}{n-m}\sum_{t=m+1}^n \varphi(X_{t-m:t}^{(r)})$
    \ENDFOR
    \STATE Fit $\hat{\pi}_{n,B} \gets \underset{f_n \in \mathcal{M}}{\arg\max} \frac{1}{B}\sum_{r=1}^{B} \log f_n(\theta^{(r)}\mid S_n(X_{1:n}^{(r)}))$
     \STATE {\bfseries Output:} Estimate $\hat{\pi}_{n,B}(\theta\mid S_n(X_{1:n}^{\mathrm{obs}}))$
  \end{algorithmic}
\end{algorithm}

\section{Theory}\label{section:theory}

In this section, we present the assumptions and main results.

\subsection{Noise Inputs}\label{subsection:noise_inputs}

We use an innovation representation to describe temporal dependence \citep{wiener_1966,rosenblatt_1961,wu_funct_dep_meas}. Let $\left(\varepsilon_i\right)_{i\in\mathbb{Z}}$ be an iid sequence of random variables, and let $\left(\varepsilon^{\ast}_i\right)_{i\in\mathbb{Z}}$, $\left(\varepsilon_i^{(r)}\right)_{i\in\mathbb{Z}}$, $r=0,1,\ldots,B$, be iid copies of $\left(\varepsilon_i\right)_{i\in\mathbb{Z}}$. For $t\in\mathbb{Z}$ and $r=0,1,\ldots,B$, define  \begin{equation}\label{eqn:noise_input_sequence_up_to_time_t}\bm{\varepsilon}_t=(\varepsilon_t,\varepsilon_{t-1},\ldots),\quad \bm{\varepsilon}_t^{(r)}=(\varepsilon_t^{(r)},\varepsilon_{t-1}^{(r)},\ldots).\end{equation}  For $t\in\mathbb{Z}$, $j\in\mathbb{N}_0$, $r=0,1,\ldots,B$, denote these sequences with the $j$-th past input replaced with a copy by \begin{equation}\label{eqn:replaced_j_past_noise_input_sequence_up_to_time_t}\begin{aligned}\tilde{\bm{\varepsilon}}_{t,j}&=\left(\varepsilon_{t},\ldots,\varepsilon_{t-j+1},\varepsilon^{\ast}_{t-j}, \varepsilon_{t-j-1},\ldots\right),\\
\tilde{\bm{\varepsilon}}_{t,j}^{(r)}&=\left(\varepsilon_{t}^{(r)},\ldots,\varepsilon_{t-j+1}^{(r)},\varepsilon^{\ast}_{t-j}, \varepsilon_{t-j-1}^{(r)},\ldots\right).\end{aligned}\end{equation}

\subsection{Assumptions}\label{subsection:assumptions}

For $t\in\mathbb{N}$ and $i\in\mathbb{N}$, we introduce mappings $G_t^{(i)}$ from $\mathbb{R}^{\infty}\times \Theta$ to $\mathbb{R}^d$ that represent the generative model. The index $n\in\mathbb{N}$ links the sample size and approximation level.
\begin{assumption}\label{asmpt:algorithmic_dynamic_model_discrete_time} The sequences of length $n\in\mathbb{N}$ are  \begin{align}\label{eqn:algorithmic_dynamic_model_discrete_time} X_{t}^{\mathrm{obs}}&=G_t^{(n)}(\bm{\varepsilon}_{t}^{(0)},\theta_0), \  X_{t}^{(r)}(\theta)=G_t^{(n)}(\bm{\varepsilon}_{t}^{(r)},\theta), \end{align} for all $t=1,\ldots,n$, $\theta\in\Theta$, and $r=1,\ldots,B$.
\end{assumption}

The next assumption is that the mappings for the approximations converge to the limiting mappings.

\begin{assumption}\label{asmpt:limiting_mapping_discrete_time} For each $t\in\mathbb{N}$, there exists a measurable map $G_t:\mathbb{R}^{\infty}\times \Theta\xrightarrow[]{}\mathbb{R}^d$, s.t., for some $q>2$, as $n\xrightarrow[]{}\infty$,   \begin{align*}\underset{\theta\in\Theta}{\sup} \left( \frac{1}{n}\sum_{t=1}^n \left\|G_t^{(n)}(\bm{\varepsilon}_0,\theta)-G_t(\bm{\varepsilon}_0,\theta)\right\|_{\mathcal{L}^q(\theta)}\right)\xrightarrow[]{} 0. \end{align*}\end{assumption}

Let the distribution of $\left[G_{t-m}(\bm{\varepsilon}_{-m},\theta),\ldots,G_t(\bm{\varepsilon}_{0},\theta)\right]^{\top}$ be denoted by $P_{\theta,t,m}$. The next assumption imposes asymptotic stationarity. Denote $\bar{P}_{\theta,n}=\frac{1}{n-m}\sum_{t=m+1}^n P_{\theta,t,m}$.
\begin{assumption}\label{asmpt:weak_convergence_of_time_avg_of_distributions_discrete_time} 
    For all $\theta \in \Theta$, there exists a limiting distribution $P_{\theta}$, such that $$\underset{\theta\in\Theta}{\sup}  \ \left|
\begin{aligned}\bar{P}_{\theta,n}\phi - P_{\theta}\phi\end{aligned} \right|\xrightarrow[]{} 0,\quad n\xrightarrow[]{}\infty,$$ for all continuous, bounded functions $\phi:\mathbb{R}^{(m+1)\times d}\xrightarrow[]{}\mathbb{R}$.
    
\end{assumption}

Next, we introduce the mappings $\Phi:\Theta\xrightarrow[]{}\mathbb{R}^k$, given by \begin{equation}\label{eqn:limiting_expectation_random_features_discrete_time} \Phi(\theta)=\int_{\mathbb{R}^{(m+1)\times d}}\varphi(x) \, dP_{\theta}(x), \quad \theta \in \Theta.\end{equation} Under Assumptions~\ref{asmpt:algorithmic_dynamic_model_discrete_time}-\ref{asmpt:weak_convergence_of_time_avg_of_distributions_discrete_time}, the (limiting) expectation values of the random features coincide with $\Phi$.

%%% Note: we're just going to assume that the parameter space is a Cartesian product of closed bounded intervals rather than assuming semialgebraic set, etc.... 

Let $\psi_{\theta}(\cdot)$ be the characteristic function of $P_{\theta}$.
\begin{assumption}\label{asmpt:statistical_model_discrete_time}
The following conditions hold:
{\renewcommand{\labelenumi}{(\alph{enumi})}\begin{enumerate} 
    \item $\Theta=\prod_{j=1}^p [a_j,b_j]$ for constants $a_j<b_j$, $j\in [p]$.    \item The map $\theta\mapsto P_{\theta}$ is injective.  
    \item The real and imaginary parts of the map $(\theta,\omega)\mapsto \psi_{\theta}(\omega)$ each admit a jointly real analytic extension to $O \times \mathbb{R}^{(m+1)\times d}$, where $O$ is open and $\Theta \subset O\subset \mathbb{R}^p$. % i.e., $O$ $is an open neighborhood of $\Theta$ 
\end{enumerate}}    
\end{assumption}
%%% Note: the first conditional can be weakened but its ok, dont want to give the details here...

The next result, adapted from \citet{random_features_estimation_paper}, is used to establish identifiability.
\begin{theorem}\label{thm:injective_discrete_time}
If Assumptions~\ref{asmpt:algorithmic_dynamic_model_discrete_time}-\ref{asmpt:statistical_model_discrete_time} hold, then, with probability one over the draw of parameters $(\vartheta_1,\ldots,\vartheta_k)$ for the $k\geq 2p+1$ features $(\varphi_1,\ldots,\varphi_k)$, the map $\theta\mapsto\Phi(\theta)$ is injective.
\end{theorem}

We assume that the parameters we have drawn for the random features yield a mapping $\theta\mapsto\Phi(\theta)$ that is injective.
\begin{assumption}\label{asmpt:injective_discrete_time}   The random parameters $(\vartheta_1,\ldots,\vartheta_k)$ we have drawn for the $k\geq 2p+1$ random Fourier features $(\varphi_1,\ldots,\varphi_k)$ yield a function $\theta\mapsto\Phi(\theta)$ that is injective. 
\end{assumption}

%%% Note: we don't need this assumption anymore

% The next condition is used to establish consistency.
% \begin{assumption}\label{asmpt:stochastic_equicontinuity_discrete_time} For all $\epsilon >0$, there exists an $\eta_G =\eta_G(\epsilon)>0$ such that $$\underset{i \in\mathbb{N}}{\sup} \ \underset{t\in\mathbb{N}}{\sup} \ \mathbb{E} \underset{ \left\|\theta-\theta'\right\|\leq \epsilon}{\underset{\theta,\theta'\in\Theta}{\sup}} \left\| G_{t}^{(i)}(\bm{\varepsilon}_{0},\theta) -  G_{t}^{(i)}(\bm{\varepsilon}_{0},\theta')\right\|< \eta_G,$$ with $\eta_G \xrightarrow[]{} 0$ as $\epsilon \xrightarrow[]{} 0$, where $\mathbb{E}(\cdot)$ is w.r.t.\ the law of $\bm{\varepsilon}_{0}$. \end{assumption}

Recall $\tilde{\bm{\varepsilon}}_{0,j}$ from~\eqref{eqn:replaced_j_past_noise_input_sequence_up_to_time_t}. The next assumption imposes a uniform moment bound and a uniform polynomial decay of the physical dependence measure from~\citet{wu_funct_dep_meas}. 
\begin{assumption}\label{asmpt:temporal_dependence_discrete_time}
    There exist $\Psi>0$, $\beta>2$ such that, for some $q > 2$ satisfying $\beta q > p$ and all $j\in\mathbb{N}_0$, \begin{align*}  \underset{i,\theta,t}{\sup} \   \left\|G_t^{(i)}(\bm{\varepsilon}_{0},\theta) - G_t^{(i)}(\tilde{\bm{\varepsilon}}_{0,j},\theta)\right\|_{\mathcal{L}^q(\theta)}   & \leq   \Psi (j \lor 1)^{-\beta},\\    \underset{i,\theta,t}{\sup} \   \left\|G_t^{(i)}(\bm{\varepsilon}_{0},\theta)\right\|_{\mathcal{L}^q(\theta)}  &\leq  \Psi. \end{align*} 
\end{assumption} 

We assume high-level regularity conditions on $\pi$ and $\pi_n$.
\begin{assumption}\label{asmpt:ref_dens} The reference density $\pi(\theta)$ is continuous and is positive on $\Theta$, and $\pi_n(\theta\mid s)$ exists for all $n\in\mathbb{N}$.
\end{assumption}

Next, we impose a smoothness condition on the distributions in terms of the Kullback--Leibler (KL) divergence.
\begin{assumption}\label{asmpt:kl_smoothness}
For all $\theta_0\in\Theta$,
there exist constants $\delta=\delta(\theta_0)>0$ and $C=C(\theta_0,\delta) >0$ such that,
for all $n\in\mathbb{N}$ and $\theta\in\Theta$ with $\|\theta-\theta_0\|\leq \delta$, we have
$$D_{\mathrm{KL}}(P_{\theta_0,n} \| P_{\theta,n})
\leq C  n \|\theta-\theta_0\|^2.
$$
\end{assumption}

We avoid detailed assumptions for a specific estimator, and simply make a high-level consistency assumption.
\begin{assumption}\label{asmpt:consistency_conditional_density_estimator} Assume that the support of $\hat{\pi}_{n,B}(\cdot \mid s)$ is $\Theta$ for all $s$, and that, for every $\theta_0\in\Theta$ and $n\in\mathbb{N}$, as $B\xrightarrow[]{}\infty$, $$\mathbb{E}
\int_{\Theta}
\left|
\hat{\pi}_{n,B}(\theta\mid S_n(X_{1:n}^{\mathrm{obs}}))
-
\pi_n(\theta\mid S_n(X_{1:n}^{\mathrm{obs}}))
\right|d\theta
\xrightarrow[]{} 0,$$ where the expectation is w.r.t.\ the distribution $P_{\theta_0,n}$ of the observations and the distribution of the simulations induced by the reference density $\pi(\theta)$ and the model $P_{\theta,n}$. 
\end{assumption}

\subsection{Results}\label{subsection:results_for_estimator}

Our first result is the concentration of the conditional density estimator around the true parameter $\theta_0$ as the sample size and number of simulations grow. Denote the complement of the ball of radius $\epsilon$ around $\theta_0$ by \begin{equation}\label{eqn:def_A}\overline{\Theta}_{\epsilon}=\{\theta\in\Theta:\left\|\theta-\theta_0\right\|>\epsilon\}.\end{equation}

\begin{theorem}\label{thm:concentration_of_cond_density_estimator}
    If Assumptions~\ref{asmpt:algorithmic_dynamic_model_discrete_time}-\ref{asmpt:consistency_conditional_density_estimator} hold, then, for any fixed $k\geq 2p+1$ and $\theta_0\in\Theta$, there exists a sequence $B_n\to\infty$ such that, for all $\epsilon>0$, as $n\xrightarrow[]{}\infty$, $$\mathbb{E}\left[\int_{\overline{\Theta}_{\epsilon}} \hat{\pi}_{n,B_n}(\theta \mid S_n(X_{1:n}^{\mathrm{obs}}))  d\theta \right] \xrightarrow[]{} 0,$$ where the expectation is w.r.t.\ the distribution $P_{\theta_0,n}$ of the observations and the distribution of the simulations induced by the reference density $\pi(\theta)$ and the model $P_{\theta,n}$.
\end{theorem}
%%% Note: expectation is with respect to the distribution P_{\theta_0,n} \otimes P_{B_n, n} where P_{\theta_0,n} is the distribution of X_{1:n} and P_{B_n, n} is the product distribution of distributions based on the simulation design (ie induced by the reference density)
%%% Note: not considering k --> infinity because we want to avoid confusing people

Denote by $\hat{\Pi}_{n,B_n}$ the conditional distribution corresponding to the estimated conditional density $\hat{\pi}_{n,B_n}(\theta \mid S_n(X_{1:n}^{\mathrm{obs}}))$, and denote by $\Pi_n^X$ the conditional distribution corresponding to the conditional density $\pi_n^X(\theta\mid X_{1:n}^{\mathrm{obs}})$.

Our second result is the convergence of $\hat{\Pi}_{n,B_n}$ to the Dirac measure $\delta_{\theta_0}$ and $\Pi_n^X$ in Wasserstein-1 distance $W_1$.

\begin{corollary}\label{cor:conv_wass_cond_distr}
    Suppose Assumptions~\ref{asmpt:algorithmic_dynamic_model_discrete_time}-\ref{asmpt:consistency_conditional_density_estimator} hold. Then, for any fixed $k\geq 2p+1$, $\theta_0\in\Theta$, there exists a sequence $B_n\to\infty$ such that, as $n\xrightarrow[]{}\infty$, $$\mathbb{E}\left[W_1\left(\hat{\Pi}_{n,B_n},\delta_{\theta_0}\right)\right]\xrightarrow[]{} 0,$$ and if for every $\eta > 0$ and $\theta_0\in\Theta$, we have $\mathbb{E}\left[\Pi_n^X(\{\theta \in \Theta : \left\|\theta-\theta_0\right\|>\eta\})\right]\xrightarrow[]{} 0$, then also    $$\mathbb{E}\left[W_1\left(\hat{\Pi}_{n,B_n},\Pi_n^X\right)\right] \xrightarrow[]{} 0,$$ where the expectation is w.r.t.\ the distribution $P_{\theta_0,n}$ of the observations and the distribution of the simulations induced by $\pi(\theta)$ and the model $P_{\theta,n}$.
\end{corollary}

%%% Note: and \Rightarrow denotes weak convergence of measures

\section{Experiments}\label{section:experiments}

We implement our sketching-based approach and all baselines using the \texttt{sbi} Python package \citep{tejerocantero2020sbitoolkitsimulationbased,boelts2025sbireloadedtoolkitsimulationbased} to ensure a fair comparison. For each method, we use a neural spline flow density estimator via the \texttt{posterior_nn} method. We compare our approach with a causal convolutional neural network (CNN) and a transformer \citep{Vaswani_et_al_2017} embedding network via the \texttt{CausalCNNEmbedding} and \texttt{TransformerEmbedding} methods, respectively, which are the two methods currently recommended for time series by the \texttt{sbi} package. 
We consider the data generating processes (DGPs) below with sample sizes $n\in\{250,500,1000\}$ and a fixed budget of $B=10000$ simulations.
%%% Note: the recommendation from the sbi package is from the following page (as of October 2026): https://sbi.readthedocs.io/en/stable/how_to_guide/20_time_series_embedding.html 

\begin{figure}[h!]
    \centering
    \includegraphics[width=\linewidth]
    {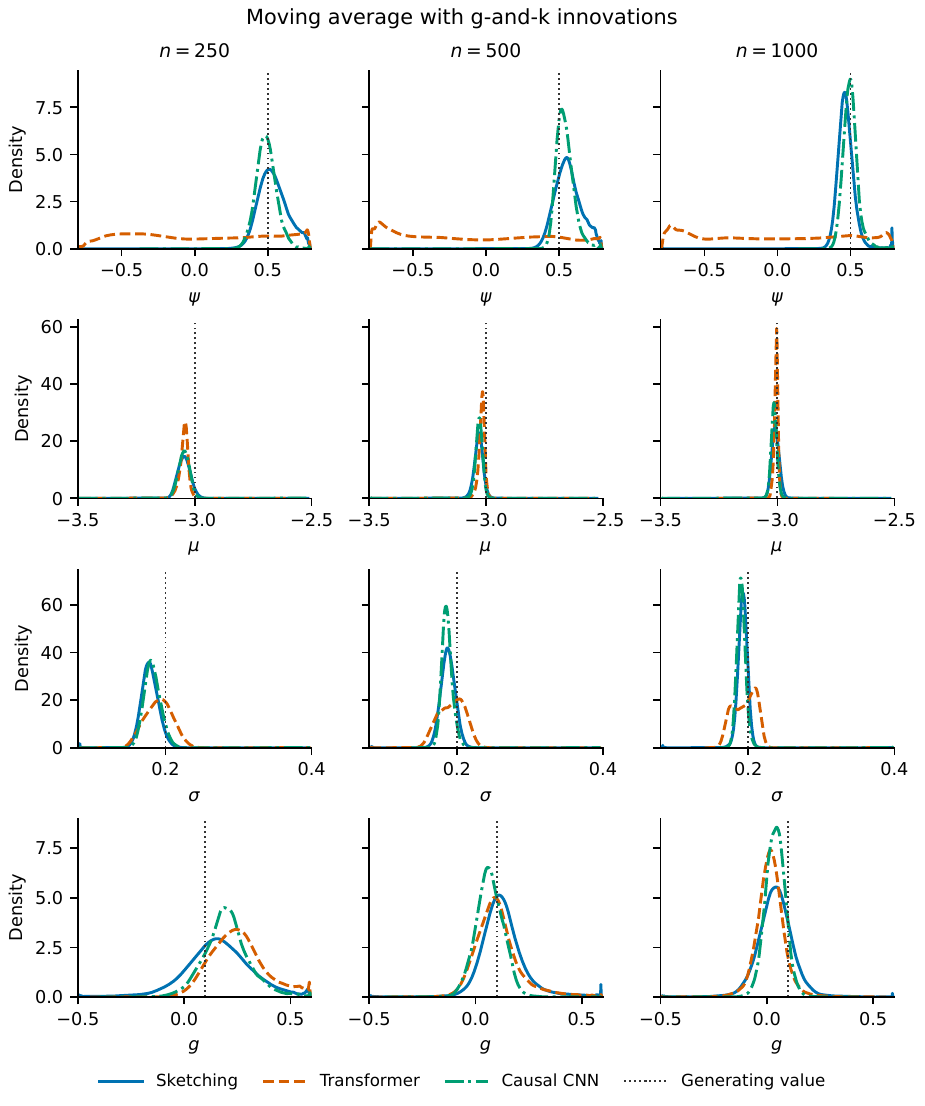}

    \caption{\textbf{Moving average with g-and-k innovations.} Marginal posterior densities from each method at different sample sizes.}
    \label{fig:gk_ma_posterior_main}
\end{figure}

\textbf{DGP 1: Moving Average Process.} We observe
\begin{equation*}
X_t =  \mu+ \psi \varepsilon_{t-1} + \varepsilon_{t}, \quad t=1,\ldots,n,
\end{equation*} where $(\varepsilon_{t})_t$ are sampled iid from the intractable g-and-k distribution~\citep{gk_MLE}. We fix the location, kurtosis, and $c$ components, respectively, to $0$, $0.1$, and $0.8$. The scale $\sigma$ and skewness $g$ are unknown. We aim to estimate $\psi_0=0.5$, $\mu_0=-3$, $\sigma_0=0.2$, $g_0=0.1$. See Figure~\ref{fig:gk_ma_posterior_main} for the results.

\begin{figure}[h!]
    \centering
    \includegraphics[width=\linewidth]
    {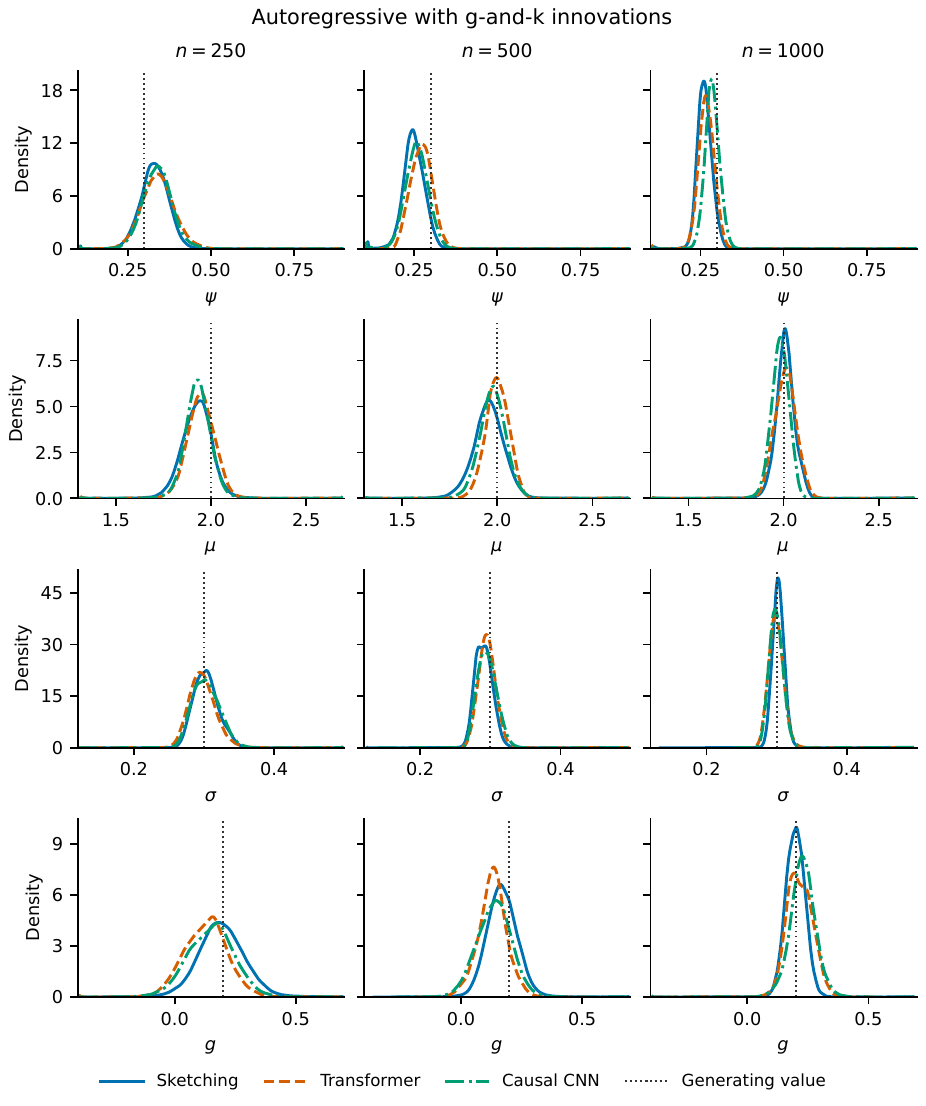}

    \caption{\textbf{Autoregressive with g-and-k innovations.} Marginal posterior densities from each method at different sample sizes.}
    \label{fig:gk_ar_posterior_main}
\end{figure}

\textbf{DGP 2: Autoregressive Process.} We observe \begin{align*}X_t &= X_{t-1} + \psi (\mu - X_{t-1}) + \varepsilon_t,\quad t=1,\ldots,n, \end{align*} where $(\varepsilon_{t})_t$ are sampled iid from the g-and-k distribution and the initial value $X_0=3$ is known. We set the location, kurtosis, and $c$ components, respectively, to $0$, $0.2$, and $0.8$, and the scale $\sigma$ and skewness $g$ are unknown. We aim to estimate $\psi_0=0.3$, $\mu_0 = 2$, $\sigma_0=0.3$, $g_0=0.2$. See Figure~\ref{fig:gk_ar_posterior_main} for the results.

\begin{figure}[h!]
    \centering
    \includegraphics[width=\linewidth]
    {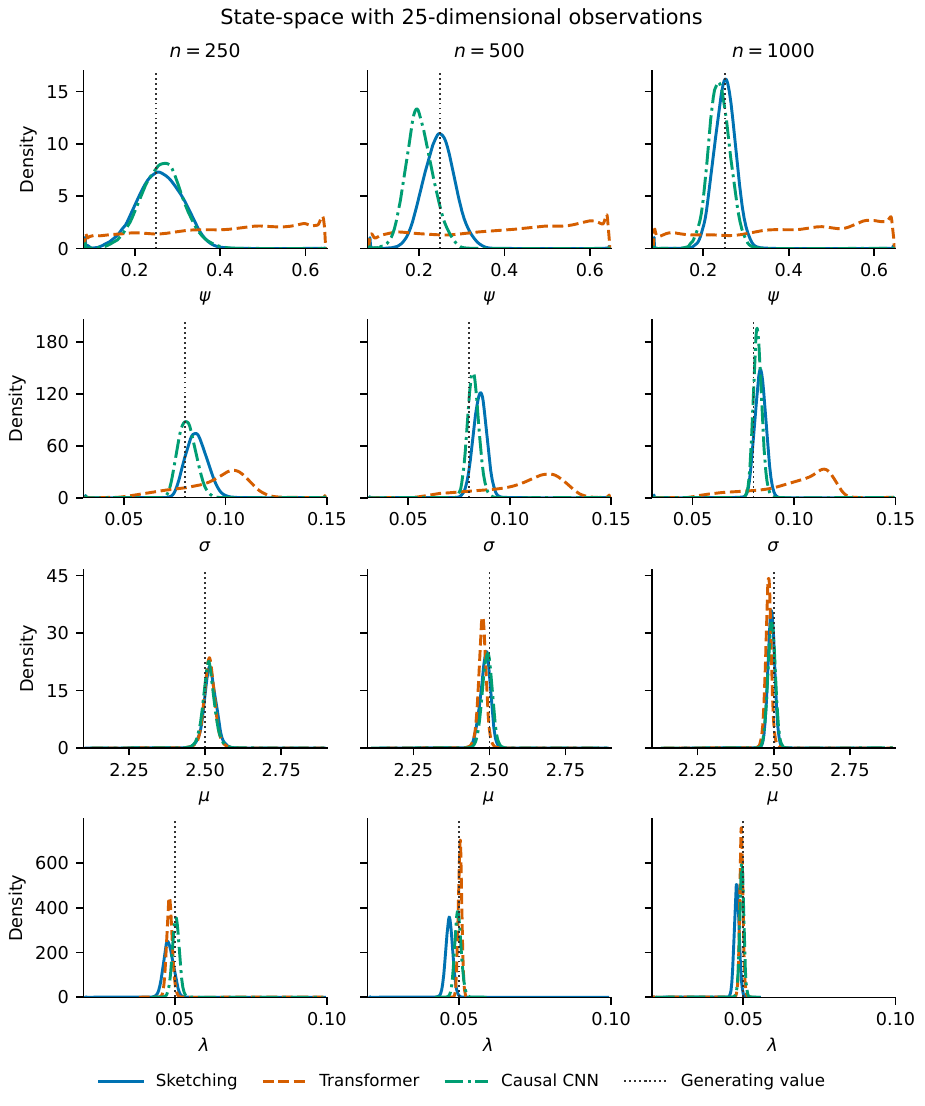}

    \caption{\textbf{State-space with 25-dimensional observations.} Marginal posterior densities from each method at different sample sizes.}
    \label{fig:state_space_posterior_main}
\end{figure}

\textbf{DGP 3: State-Space Model.} The latent state is given by 
$$Z_{t} = Z_{t-1} + \psi (\mu - Z_{t-1}) + \sigma \varepsilon_t,$$ with $\varepsilon_t \overset{\mathrm{iid}}{\sim} N(0,1)$. For $j=1,\ldots,d\equiv 25$, we observe \begin{align*}X_{t}^{(j)} &= Z_{t} + \lambda \xi_{t}^{(j)}, \quad t=1,\ldots,n, 
\end{align*} where $\xi_{t}^{(j)} \overset{\mathrm{iid}}{\sim} N(0,1)$. We aim to estimate $\psi_0=0.25$, $\sigma_0=0.08$, $\mu_0 = 2.5$, $\lambda_0=0.05$. See Figure~\ref{fig:state_space_posterior_main} for the results. 
%%% Note: To situate this model within our theoretical framework, for the process $Z_{t}$, we let the number of noise inputs in the burn-in period grow with the sample size. 

\begin{figure}[h!]
    \centering
    \includegraphics[width=\linewidth]
    {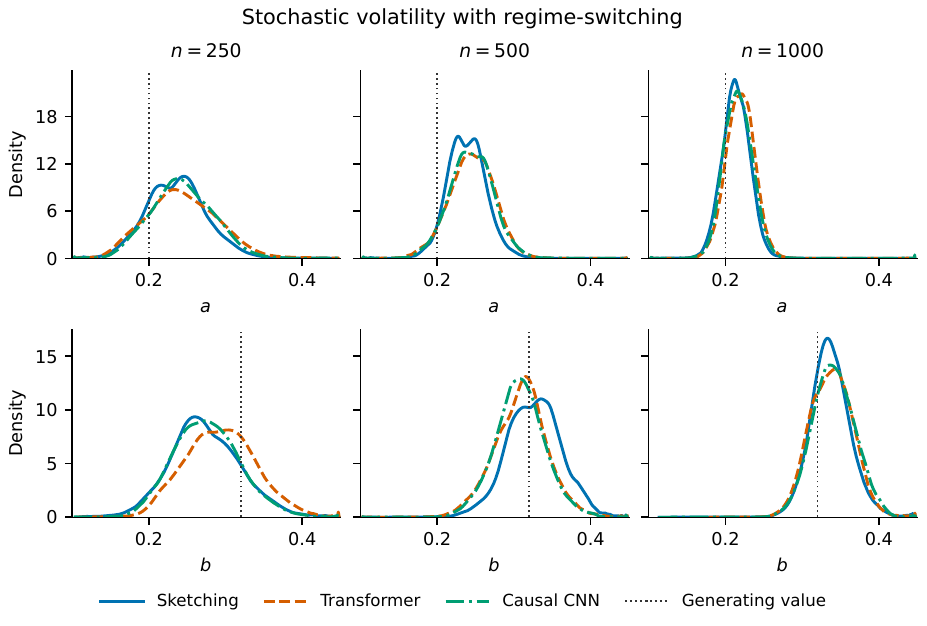}

    \caption{\textbf{Stochastic volatility with regime-switching.} Marginal posterior densities from each method at different sample sizes.}
    \label{fig:switching_posterior_main}
\end{figure}

\textbf{DGP 4: Regime-Switching Stochastic Volatility Model.} The latent volatility process is given by $$Z_t = \psi Z_{t-1} + \rho \sqrt{1-\psi^2}\eta_t,$$ where $\eta_t\overset{\mathrm{iid}}{\sim}N(0,1)$ and $Z_0\sim N(0,\rho^2)$. The latent state $S_t$ takes on values $\{-1,1\}$ and has transition probabilities 
\begin{align*}P(S_t=1 \mid S_{t-1}=-1)&=a,\\  
P(S_t=-1 \mid S_{t-1}=1)&=b,
\end{align*}
and initial condition $P(S_0=1)=a/(a+b) 
$. We observe \begin{align*}
    X_t = S_t + \sigma \exp(Z_t / 2) \varepsilon_t, \quad t=1,\ldots,n,
\end{align*} where $\varepsilon_t\overset{\mathrm{iid}}{\sim}N(0,1)$. We know $\psi_0=0.75$, $\rho_0=0.5$, and $\sigma_0=0.45$, and we aim to estimate $a_0=0.2$ and $b_0=0.32$. See Figure~\ref{fig:switching_posterior_main} for the results. Note that this example violates the real analyticity assumption from Assumption~\ref{asmpt:statistical_model_discrete_time}, but we include it to show that our method still performs well outside of the theoretical framework we used in this paper.

\begin{figure}[h!]
    \centering
    \includegraphics[width=\linewidth]
    {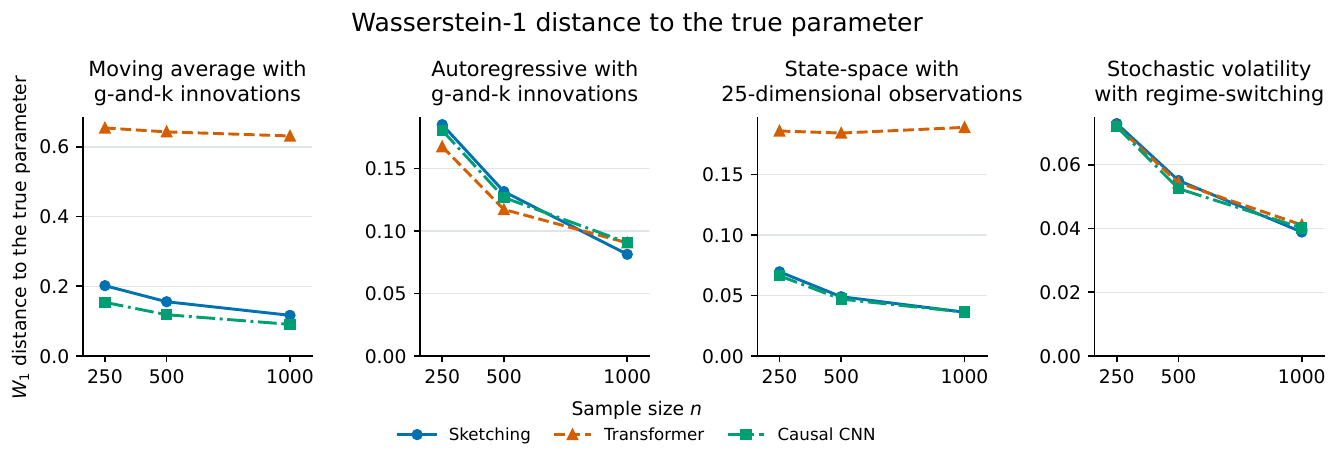}

    \caption{\textbf{Comparison of posterior density quality}. Wasserstein-1 distance to $\delta_{\theta_0}$ at different sample sizes, averaged over $1000$ trajectories simulated under $P_{\theta_0,n}$.}
    \label{fig:wasserstein_to_truth}
\end{figure}

\begin{figure}[h!]
    \centering
    \includegraphics[width=\linewidth]
    {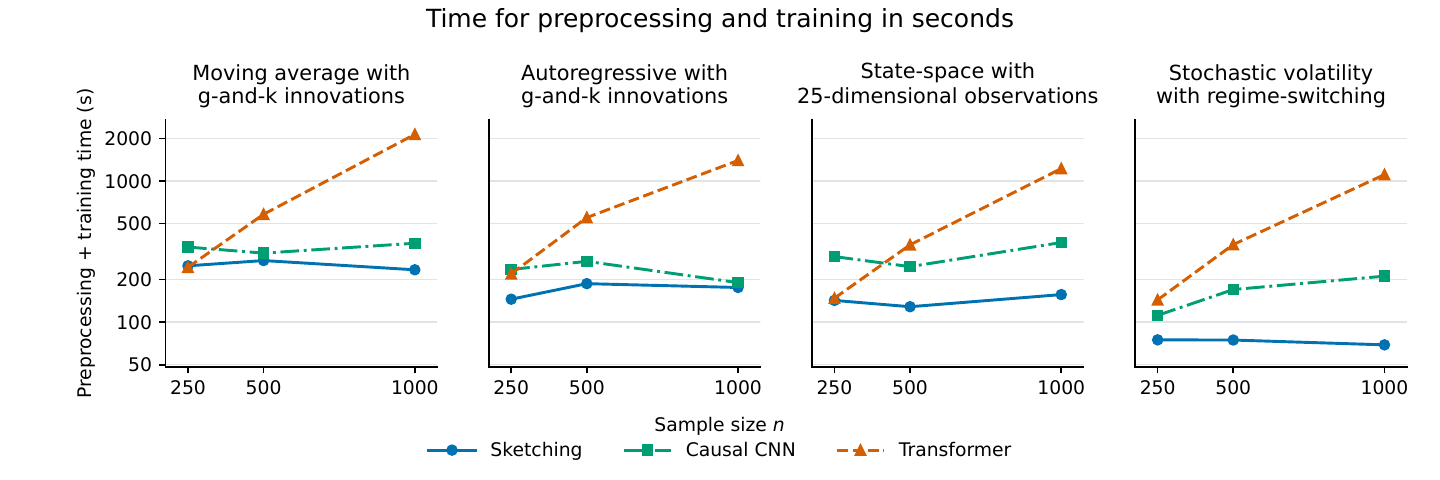}

    \caption{\textbf{Comparison of total computation cost.} Preprocessing and training time at different sample sizes.}
    \label{fig:sbi_preprocessing_training_time}
\end{figure}

\begin{figure}[h!]
    \centering
    \includegraphics[width=\linewidth]
    {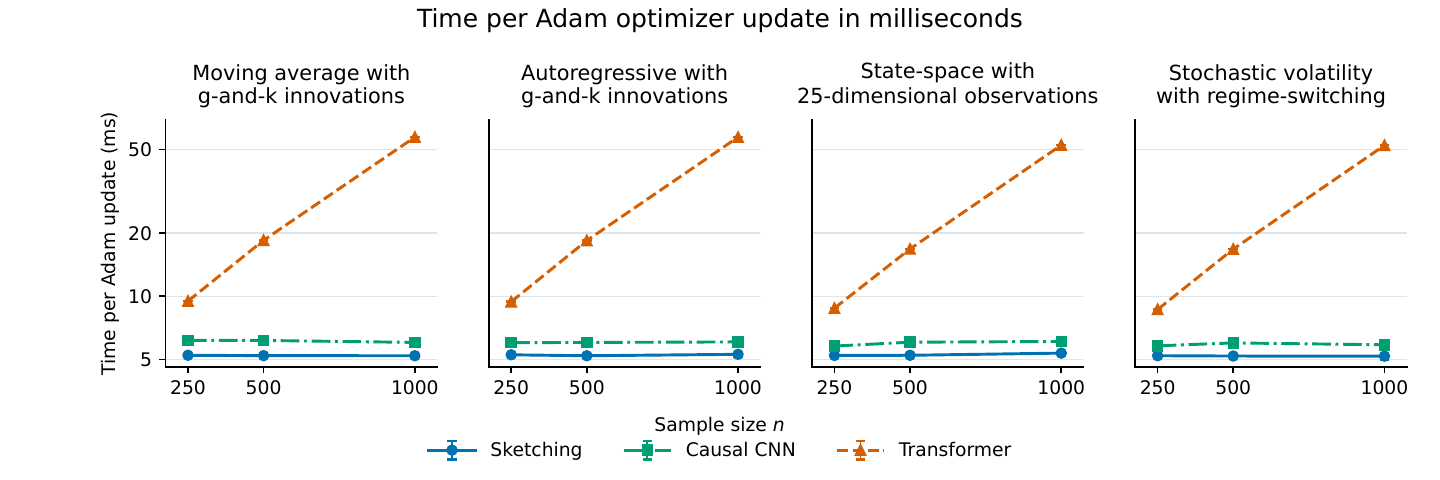}

    \caption{\textbf{Comparison of optimization speed.} Adam optimizer update time at different sample sizes.}
    \label{fig:sbi_adam_update_time}
\end{figure}

\textbf{Interpretation of Results.} We find that our method performs comparably (see Figure~\ref{fig:wasserstein_to_truth}) to the state-of-the-art \texttt{CausalCNNEmbedding} embedding network, while being more computationally efficient (see Figures~\ref{fig:sbi_preprocessing_training_time} and~\ref{fig:sbi_adam_update_time}). This makes sense because our approach can be viewed as a \textit{randomized} 1-dimensional convolutional neural network (1-D CNN). Specifically, a 1-D CNN where the convolutional filters are chosen to be \textit{random} Fourier features with $m$ lags (i.e., no learned weights) and when a time-average pooling operation is used.

\section{Conclusion and Future Work}\label{section:conclusion_future_work}

We propose and study a method for neural SBI based on \textit{randomized} summary statistics. Our method fills a gap in the neural SBI literature for situations when the model's complexity makes it challenging to manually design informative summary statistics, and when the high dimensionality of the data makes learning good summary statistics too computationally expensive. In particular, we show that our approach is agnostic to the sample size $n$ and observation dimension $d$, and that the number of sketches sufficient for posterior consistency scales linearly in the parameter dimension $p$.

Using the \texttt{sbi} Python package (to ensure an apples-to-apples comparison), we compare our method with state-of-the-art embedding networks. Our results imply that it is not necessary to learn summary statistics for NPE because $k\geq 2p+1$ random Fourier features perform just as well as state-of-the-art embedding networks \textit{at a reduced computational cost}. We emphasize that, although randomized convolutional filters are not a new idea, their usage is typically justified in an ad-hoc way. As far as we know, we are the first to \textit{theoretically analyze} posterior consistency in neural SBI in the context of randomized summary statistics based on time-averages of random Fourier features. Specifically, we show that $k\geq 2p+1$ random Fourier features are sufficient for posterior consistency. Our theoretical analysis may be of independent interest for analyzing other randomized neural network architectures since random \textit{Fourier} features in particular are not needed. 

There are several promising directions for future work. First, our ideas can likely be extended to other types of data and dependencies, e.g., network and spatiotemporal. Second, it would be interesting to consider other types of sketching for different data types. Third, it would be worthwhile to consider how randomized summary statistics could be used in other areas of SBI. Lastly, connecting our ideas to recent work on the intrinsic information rates of summary statistics \cite{bansal2026intrinsicinformationratessummaries} would be a promising next step.

%\section*{Acknowledgements}

\section*{Impact Statement}
This paper presents work whose goals are to advance the fields of statistics and machine
learning. There are many potential societal consequences of our work, none of which we feel must be specifically highlighted here.

\subsection*{AI use statement}

% following this: https://iclr.cc/Conferences/2027/AIPolicyForAuthors

Starting from core ideas of the authors, we acknowledge that generative AI was used to: identify and search for relevant literature, brainstorm ideas involving the conceptual frameworks, suggest edits to the paper to improve prose, suggest edits to the proofs to check correctness, shorten length, and improve readability, suggest edits to software code for the method, experiments, and figures, suggest experimental parameters for experiments, identify possible gaps in the paper, and provide feedback on the experiments. 

Generative AI was not used to: draft parts of the paper, format references, formulate the presented mathematical claims, provide the critical ingredients for proving mathematical claims,  suggest a structure for the paper, assist with translation, interpret results, generate synthetic data sets, or propose a title or keywords for the paper. 

Research subtasks not applicable to this work: propose or refine hypotheses, clean and reformat dataset, support qualitative and thematic data analysis, formulate questions for surveys or interviews, create artifacts,  transcribe recordings of research material.

The authors have reviewed all content suggested by generative AI and produced with the aid of generative AI. For example, the authors have reviewed all literature found by generative AI and checked the correctness of any suggestions made by generative AI. We take responsibility for the final content of this work, including text, claims or artifacts produced with the aid of generative AI.

\bibliography{neural_posterior_references}
\bibliographystyle{icml2026}

% APPENDIX

\newpage
\appendix
\onecolumn

\let\addcontentsline\savedaddcontentsline
\setcounter{tocdepth}{2}
\tableofcontents

\section{Preliminaries}

We state several useful results for working with time series of random features. First, we establish the Lipschitz smoothness of the random Fourier features.

\begin{lemma}\label{lma:lipschitz_features} The function $\varphi=(\varphi_1,\ldots,\varphi_k)$ from~\eqref{eqn:all_k_random_Fourier_features_order_m} is Lipschitz with Lipschitz constant  \begin{equation}\label{eqn:lipschitz_constant_rff}L_{\varphi}=\left(\sum_{i=1}^k\sum_{j=1}^{m+1}\left\|\Omega_{i,j}\right\|^2\right)^{1/2}.\end{equation}
\end{lemma}
\begin{proof}
    See the proof of Lemma 1 in \citet{random_features_estimation_paper}.
\end{proof}

Next, we introduce some useful notation. For $t\geq m+1$, $i\in\mathbb{N}$, $j\in\mathbb{N}_0$, $r=0,1,\ldots,B$, define
\begin{align*}Z_{t}^{(i,r)}(\theta)&=\left[G_{t-m}^{(i)}(\bm{\varepsilon}_{-m}^{(r)},\theta),\ldots,G_t^{(i)}(\bm{\varepsilon}_0^{(r)},\theta)\right]^{\top}, 
\\ 
\tilde{Z}_{t,j}^{(i,r)}(\theta) &=\left[G_{t-m}^{(i)}(\bar{\bm{\varepsilon}}_{-m,j}^{(r)},\theta),\ldots,G_t^{(i)}(\bar{\bm{\varepsilon}}_{0,j}^{(r)},\theta)\right]^{\top},\end{align*}
where for $\ell=0,\ldots,m$ we define $$\bar{\bm{\varepsilon}}_{-\ell,j}^{(r)}=\begin{cases} \tilde{\bm{\varepsilon}}_{-\ell,j-\ell}^{(r)} \text{ for } \ell \leq j,
\\ \bm{\varepsilon}_{-\ell}^{(r)} \text{ for } \ell > j,
\end{cases}$$
where $\tilde{\bm{\varepsilon}}_{-\ell,j-\ell}^{(r)}$ is from~\eqref{eqn:replaced_j_past_noise_input_sequence_up_to_time_t}.

The following result establishes that the time series of random Fourier features has decaying temporal dependence in terms of the physical dependence measure \citep{wu_funct_dep_meas}.

\begin{lemma}\label{lma:temporal_dependence_of_random_features_discrete_time} Under Assumptions~\ref{asmpt:algorithmic_dynamic_model_discrete_time} and~\ref{asmpt:temporal_dependence_discrete_time}, there exist $\Psi^{\varphi}>0$, $\beta^{\varphi}>2$ such that, for some $q^{\varphi} > 2$ and all $j\in\mathbb{N}_0$,   \begin{align*}  \underset{i \in\mathbb{N}}{\sup} \ \underset{\theta\in\Theta}{\sup} \ \underset{t\geq m+1}{\sup}   \  \underset{0\leq r \leq B}{\max} \   \left\|\varphi\left(Z_{t}^{(i,r)}(\theta)\right) - \varphi\left(
\tilde{Z}_{t,j}^{(i,r)}(\theta)\right)\right\|_{\mathcal{L}^{q^{\varphi}}(\theta)}   & \leq   \Psi^{\varphi} (j \lor 1)^{-\beta^{\varphi}},\\  \underset{i \in\mathbb{N}}{\sup} \ \underset{\theta\in\Theta}{\sup} \ \underset{t\geq m+1}{\sup} \ \underset{0\leq r \leq B}{\max} \ \left\|\varphi\left(Z_{t}^{(i,r)}(\theta)\right)\right\|_{\mathcal{L}^{q^{\varphi}}(\theta)}  &\leq  \Psi^{\varphi}. \end{align*} 
\end{lemma}
\begin{proof}
    See the proof of Lemma 3 in \citet{random_features_estimation_paper}.
\end{proof}

The next result is Theorem 3.2 from \citet{seq_gauss_approx2022} in our notation.
\begin{lemma}\label{lma:lp_lln_discrete_time} 
For $r=0,1,\ldots,B$ and $\theta\in\Theta$, suppose a time series $X_t^{(r)}(\theta)$, $t=1,\ldots,n$, is generated as $$X_t^{(r)}(\theta)=G_{t}^{(n)}(\bm{\varepsilon}_{t},\theta),$$ for some input sequence $\bm{\varepsilon}_{t}$. Suppose there exists a constant $q\geq 2$, such that $$\underset{i\in\mathbb{N}}{\sup} \ \underset{\theta\in\Theta}{\sup} \  \underset{t\in\mathbb{N}}{\sup} \ \underset{0\leq r \leq B}{\max} \  \left\|G_t^{(i)}(\bm{\varepsilon}_{0},\theta)\right\|_{\mathcal{L}^q(\theta)}<\infty.$$ For some $2\leq h \leq q < \infty$, define the physical dependence measure at time $t$ and sample size $n$ as $$\eta_{t,j,q,h}^{(n)}(\theta)=  \mathbb{E}_{\theta}\left(\left\|G_t^{(n)}(\bm{\varepsilon}_{t},\theta) - G_t^{(n)}(\tilde{\bm{\varepsilon}}_{t,j},\theta)\right\|_h^q\right)^{1/q}.$$ There exists a constant $C = C(q,h)$, such that for all $r=0,1,\ldots,B$, $\theta\in\Theta$, and $n \in \mathbb{N}$, we have
\begin{align*}&
 \left(
\mathbb{E}_{\theta} \left[\max_{\ell \leq n}
\left\|
\sum_{t=1}^{\ell} \left(X_t^{(r)}(\theta) - \mathbb{E}_{\theta}\left[X_t^{(r)}(\theta)\right]\right)
\right\|_h^q\right]
\right)^{\frac{1}{q}}\\
&=
 \left(
\mathbb{E}_{\theta} \left[\max_{\ell \leq n}
\left\|
\sum_{t=1}^{\ell} \left(G_t^{(n)}(\bm{\varepsilon}_{t},\theta) - \mathbb{E}_{\theta}\left[G_t^{(n)}(\bm{\varepsilon}_{t},\theta)\right]\right)
\right\|_h^q\right]
\right)^{\frac{1}{q}}\\
&\leq
C n^{\frac{1}{2} - \frac{1}{q}}
\sum_{j=0}^{\infty}
\left(
\sum_{t=1}^n (\eta_{t,j,q,h}^{(n)}(\theta))^q
\right)^{\frac{1}{q}} \\ &\leq
C n^{\frac{1}{2}}
\sum_{j=0}^{\infty}
\max_{t \leq n} \eta_{t,j,q,h}^{(n)}(\theta).
\end{align*} % and in the special case of $h = 2$, the inequality may be improved to \begin{align*}& \left(\mathbb{E}_{\theta}\left[ \max_{\ell \leq n}\left\|\sum_{t=1}^{\ell} \left(X_t^{(r)}(\theta) - \mathbb{E}_{\theta}\left[X_t^{(r)}(\theta)\right]\right)\right\|_2^q\right]\right)^{\frac{1}{q}}\\&=  \left(\mathbb{E}_{\theta}\left[ \max_{\ell \leq n}\left\|\sum_{t=1}^{\ell} \left(G_t^{(n)}(\bm{\varepsilon}_{t},\theta) - \mathbb{E}_{\theta}\left[G_t^{(n)}(\bm{\varepsilon}_{t},\theta)\right]\right)\right\|_2^q\right]\right)^{\frac{1}{q}}\\&\leq C \sum_{j=0}^{\infty} (j \wedge n)^{\frac{1}{2} - \frac{1}{q}}\left(\sum_{t=1}^n (\eta_{t,j,q,2}^{(n)}(\theta))^q \right)^{\frac{1}{q}}+ C \sum_{j=0}^n\left(\sum_{t=1}^n (\eta_{t,j,2,2}^{(n)}(\theta))^2\right)^{\frac{1}{2}}.\end{align*}
\end{lemma}
\begin{proof}
    See the proof of Lemma 4 in \citet{random_features_estimation_paper}.
\end{proof}

\section{Proof of Theorem~\ref{thm:injective_discrete_time}}\label{section:proof_thm_injective_discrete_time}

The result follows by the same arguments as the proof of Theorem 1 in \citet{random_features_estimation_paper}, replacing $k=2p+1$ by $k\geq 2p+1$. Note that the finite witness theorem (Theorem 6 in \citet{random_features_estimation_paper}, Corollary A.20 in \citet{amir_et_al_finite_witness_thm}) can be applied with any $k\geq 2p+1$; see the
proof of Proposition 3.6 in \citet{amir_et_al_finite_witness_thm}.

\section{Proof of Theorem~\ref{thm:concentration_of_cond_density_estimator}}\label{section:proof_thm_concentration}

The proof is broken into four steps.

Note that if $\overline{\Theta}_{\epsilon}=\emptyset$ for the chosen $\epsilon >0$, then the result follows trivially. Hence, for everything below, assume that $\overline{\Theta}_{\epsilon}\neq \emptyset$.

\paragraph*{Step 1.}

Choose a constant $q^{\ast} \in (\max\{2,p\}, q\beta)$, where $p$ is the parameter dimension, and $q$, $\beta$ are from Assumption~\ref{asmpt:temporal_dependence_discrete_time}. Since the random Fourier features are bounded and Lipschitz, Assumption~\ref{asmpt:temporal_dependence_discrete_time} implies that the physical dependence measure for the time series of random Fourier features satisfies
$$\eta_{t,j,q^{\ast},2}^{(n)}(\theta) \leq C_{\varphi}(j\lor 1)^{-\gamma},\quad \gamma=\beta \min(1,q/q^{\ast})>1.$$
for some $C_{\varphi}>0$, which follows by arguments analogous to those used to show Lemma~\ref{lma:temporal_dependence_of_random_features_discrete_time}. Observe that \begin{align*}
& \underset{\theta\in\Theta}{\sup} \ \mathbb{E}_{\theta}\left(\left\|S_n(X_{1:n}(\theta)) - \mathbb{E}_{\theta}\left[S_n(X_{1:n}(\theta))\right]\right\|^{q^{\ast}} \right)^{1/q^{\ast}}
\\& 
\overset{(1)}{\leq}
2 \ C \ \frac{1}{n-m} \ (n-m)^{1/2-1/q^{\ast}} \ \underset{\theta\in\Theta}{\sup}  \ \sum_{j=0}^{\infty}
\left(
\sum_{t=m+1}^n (\eta_{t,j,q^{\ast},2}^{(n)}(\theta))^{q^{\ast}}
\right)^{\frac{1}{q^{\ast}}}
\\&
\overset{(2)}{\leq}
2 \ C \ \frac{1}{n-m}  \ (n-m)^{1/2-1/q^{\ast}} \  \underset{\theta\in\Theta}{\sup}  \ \sum_{j=0}^{\infty}
(n-m)^{\frac{1}{q^{\ast}}} \underset{t\in\{m+1,\ldots,n\}}{\max} \ \eta_{t,j,q^{\ast},2}^{(n)}(\theta)
\\&
\overset{(3)}{\leq} 
2 \ C \ \frac{1}{\sqrt{n-m}} \ C_{\varphi} \ \sum_{j=0}^{\infty} (j\lor 1)^{-\gamma}
\\&
\overset{(4)}{\leq} 
2 \ C \ \frac{1}{\sqrt{n-m}} \ C_{\varphi}  K, 
\end{align*}
by (1) factoring out $\frac{1}{n-m}$ then applying the first inequality from Lemma~\ref{lma:lp_lln_discrete_time} to the time series of random features and further upper bounding with infinitely many terms in the summation, (2) upper bounding the sum by $n-m$ times the maximum and taking the $q^{\ast}$-th root, (3) by bounds on the physical dependence measures of the original time series by
Assumption~\ref{asmpt:temporal_dependence_discrete_time} and of the time series of random features shown above, and (4) because $\gamma > 1$ so the series converges. Hence, taking the $q^{\ast}$-th power of both sides, we have
$$\underset{\theta\in\Theta}{\sup} \ \mathbb{E}_{\theta}\left(\left\|S_n(X_{1:n}(\theta)) - \mathbb{E}_{\theta}\left[S_n(X_{1:n}(\theta))\right]\right\|^{q^{\ast}} \right)=O(n^{-\frac{q^{\ast}}{2}}).$$

\paragraph*{Step 2.} Second, we consider the term $\sup_{\theta\in\Theta}\left\|\mathbb{E}_{\theta}\left[S_n(X_{1:n}(\theta))\right] - \Phi(\theta)\right\|$. For each $r\in\{0,1,\ldots,B\}$, \begin{align*}&\sup_{\theta\in\Theta}\left\|\mathbb{E}_{\theta}\left[S_n(X_{1:n}(\theta))\right] - \Phi(\theta)\right\| \\
&\overset{(1)}{=} \underset{\theta\in\Theta}{\sup} \ \left\|\frac{1}{n-m}\sum_{t=m+1}^n \mathbb{E}_{\theta}\left[ \varphi\left(\left[G_{t-m}^{(n)}(\bm{\varepsilon}_{t-m}^{(r)},\theta),\ldots,G_t^{(n)}(\bm{\varepsilon}_t^{(r)},\theta)\right]^{\top}\right)\right] - \Phi(\theta)\right\|
\\&\overset{(2)}{\leq} \sup_{\theta\in\Theta} 
\left\|
\begin{aligned}
&\frac{1}{n-m}\sum_{t=m+1}^n 
\mathbb{E}_{\theta}\left[
\varphi\left(
\left[G_{t-m}^{(n)}(\bm{\varepsilon}_{t-m}^{(r)},\theta),\ldots,
G_t^{(n)}(\bm{\varepsilon}_t^{(r)},\theta)\right]^{\top}
\right)
\right] \\
&-
\frac{1}{n-m}\sum_{t=m+1}^n 
\mathbb{E}_{\theta}\left[
\varphi\left(
\left[G_{t-m}(\bm{\varepsilon}_{t-m}^{(r)},\theta),\ldots,
G_{t}(\bm{\varepsilon}_t^{(r)},\theta)\right]^{\top}
\right)
\right]
\end{aligned}
\right\| 
\\ &
+ \sup_{\theta\in\Theta}
\left\|
\frac{1}{n-m}\sum_{t=m+1}^n 
\mathbb{E}_{\theta}\left[
\varphi\left(
\left[G_{t-m}(\bm{\varepsilon}_{t-m}^{(r)},\theta),\ldots,
G_{t}(\bm{\varepsilon}_t^{(r)},\theta)\right]^{\top}
\right)
\right] - \Phi(\theta)
\right\| 
\\ 
& \overset{(3)}{=} o(1),
\end{align*}
by (1) the definition of $S_n(X_{1:n}(\theta))$ and linearity of expectation, (2) adding and subtracting the same term, the triangle inequality, and subadditivity of the supremum, (3) because both terms are $o(1)$. Specifically, the second term is $o(1)$ by upper bounding the $\ell_2$ norm $\left\|\cdot\right\|$ by the $\ell_1$ norm (i.e., over the $k$ random features) using the monotonicity of norms, subadditivity of the supremum, linearity, and then applying the weak convergence from Assumption~\ref{asmpt:weak_convergence_of_time_avg_of_distributions_discrete_time}. The first term is $o(1)$ because the approximations can be asymptotically replaced by the limiting mappings, because for all $r \in \{0,1,\ldots,B\}$, we have \begin{align*} & \underset{\theta\in\Theta}{\sup} \ 
\left\|
\begin{aligned}
&\frac{1}{n-m}\sum_{t=m+1}^n 
\mathbb{E}_{\theta}\left[
\varphi\left(
\left[G_{t-m}^{(n)}(\bm{\varepsilon}_{t-m}^{(r)},\theta),\ldots,
G_t^{(n)}(\bm{\varepsilon}_t^{(r)},\theta)\right]^{\top}
\right)
\right] \\
&-
\frac{1}{n-m}\sum_{t=m+1}^n 
\mathbb{E}_{\theta}\left[
\varphi\left(
\left[G_{t-m}(\bm{\varepsilon}_{t-m}^{(r)},\theta),\ldots,
G_{t}(\bm{\varepsilon}_t^{(r)},\theta)\right]^{\top}
\right)
\right]
\end{aligned}
\right\|
\\
&\overset{(1)}{\leq} 
\underset{\theta\in\Theta}{\sup} \left(\frac{1}{n-m}\sum_{t=m+1}^n   \left\|
\begin{aligned}
& 
\varphi\left(
\left[G_{t-m}^{(n)}(\bm{\varepsilon}_{t-m}^{(r)},\theta),\ldots,
G_t^{(n)}(\bm{\varepsilon}_t^{(r)},\theta)\right]^{\top}
\right)
 \\
& - 
\varphi\left(
\left[G_{t-m}(\bm{\varepsilon}_{t-m}^{(r)},\theta),\ldots,
G_{t}(\bm{\varepsilon}_t^{(r)},\theta)\right]^{\top}
\right)
\end{aligned}
\right\|_{\mathcal{L}^1(\theta)}\right)
\\
& \overset{(2)}{\leq}  L_{\varphi} \ \underset{\theta\in\Theta}{\sup} \left(
\frac{1}{n-m}\sum_{t=m+1}^n    \left\|
\begin{aligned}
&  
\mathrm{Vec}\left(G_{t-m}^{(n)}(\bm{\varepsilon}_{t-m}^{(r)},\theta),\ldots,
G_t^{(n)}(\bm{\varepsilon}_t^{(r)},\theta)\right)
 \\
&- \mathrm{Vec}
\left(G_{t-m}(\bm{\varepsilon}_{t-m}^{(r)},\theta),\ldots,
G_{t}(\bm{\varepsilon}_t^{(r)},\theta)\right)
\end{aligned}
\right\|_{\mathcal{L}^1(\theta)}\right)
\\
& \overset{(3)}{\leq}  L_{\varphi} \ \underset{\theta\in\Theta}{\sup} \left(
\frac{1}{n-m}\sum_{t=m+1}^n \sum_{j=0}^m    \left\|
G_{t-j}^{(n)}(\bm{\varepsilon}_{t-j}^{(r)},\theta) - 
G_{t-j}(\bm{\varepsilon}_{t-j}^{(r)},\theta)
\right\|_{\mathcal{L}^1(\theta)}\right)
\\
& \overset{(4)}{\leq}
 L_{\varphi} 
\frac{1}{1-m/n} (m+1) \ \underset{\theta\in\Theta}{\sup} \left(\frac{1}{n}\sum_{t=1}^n    \left\|G_{t}^{(n)}(\bm{\varepsilon}_{0}^{(r)},\theta) - 
G_{t}(\bm{\varepsilon}_{0}^{(r)},\theta)
\right\|_{\mathcal{L}^1(\theta)}\right)
\\
& \overset{(5)}{=}  o(1),
\end{align*}
where (1) holds by linearity of summation, the triangle inequality, linearity of expectation, and Jensen's inequality, (2) holds because the random Fourier features $\varphi$ are $L_{\varphi}$-Lipschitz with Lipschitz constant given by~\eqref{eqn:lipschitz_constant_rff}, (3) holds by the triangle inequality and Minkowski inequality, (4) follows by upper bounding the original quantity by $m+1$ times the full sum (i.e., by including more non-negative terms in the summation), factoring out $1/n$, and noting that each $(G_t^{(i)}(\bm{\varepsilon}_j,\theta))_{j\in\mathbb{Z}}$, $(G_{t}(\bm{\varepsilon}_j,\theta))_{j\in\mathbb{Z}}$ is a stationary ergodic process for each fixed $t\in\mathbb{N}$, $i\in \mathbb{N}$, and $r=0,1,\ldots,B$, which allows us to shift the noise inputs, and (5) is from applying the convergence condition from Assumption~\ref{asmpt:limiting_mapping_discrete_time}.

For every $a>0$, the previous inequalities imply that, for sufficiently large sample size $n$, \begin{equation}\label{eqn:limiting_mean_convergence}\sup_{\theta\in\Theta}\left\|\mathbb{E}_{\theta}\left[S_n(X_{1:n}(\theta))\right] - \Phi(\theta)\right\|< a/2,\end{equation} and that
\begin{align*}
&    \sup_{\theta\in\Theta}P_{\theta,n}(\left\|S_n(\theta)-\Phi(\theta)\right\|>a)
\\&
\overset{(1)}{\leq} \sup_{\theta\in\Theta} P_{\theta,n}(\left\|S_n(\theta)-\mathbb{E}_{\theta}\left[S_n(X_{1:n}(\theta))\right] \right\|
+\left\|\mathbb{E}_{\theta}\left[S_n(X_{1:n}(\theta))\right] -\Phi(\theta)\right\|>a)
\\&
\overset{(2)}{=}
\sup_{\theta\in\Theta} P_{\theta,n}(\left\|S_n(\theta)-\mathbb{E}_{\theta}\left[S_n(X_{1:n}(\theta))\right] \right\|
>a-\left\|\mathbb{E}_{\theta}\left[S_n(X_{1:n}(\theta))\right] -\Phi(\theta)\right\|)
\\&
\overset{(3)}{\leq}
\sup_{\theta\in\Theta} P_{\theta,n}(\left\|S_n(\theta)-\mathbb{E}_{\theta}\left[S_n(X_{1:n}(\theta))\right] \right\|
>a/2)
\\&
\overset{(4)}{\leq}
\left(\frac{2}{a}\right)^{q^{\ast}}
\sup_{\theta\in\Theta} \mathbb{E}_{\theta}\left\|S_n(\theta)-\mathbb{E}_{\theta}\left[S_n(X_{1:n}(\theta))\right] \right\|^{q^{\ast}}
\\&
\overset{(5)}{\leq}
\left(\frac{2}{a}\right)^{q^{\ast}}
C(n-m)^{-q^{\ast}/2}
\\&
=
O(n^{-q^{\ast}/2}),
\end{align*}
where (1) is by the triangle inequality and inclusion of events, (2) is by subtraction, (3) is by the inequality from~\eqref{eqn:limiting_mean_convergence}, (4) is by taking the $q^{\ast}$-th power on both sides and applying Markov's inequality, and (5) is by the inequality from Step 1 for some constant $C>0$.

Therefore, for every $a>0$ and sufficiently large sample size $n$, we have
\begin{equation}\label{eqn:limiting_mean_conv_Phi}\sup_{\theta\in\Theta}P_{\theta,n}(\left\|S_n(\theta)-\Phi(\theta)\right\|>a)\leq C_a (n-m)^{-q^{\ast}/2},\end{equation} for some constant $C_a>0$.

\paragraph*{Step 3.} 

\paragraph*{Step 3.1.}

Define \begin{align}\label{eqn:def_B} \rho_{\epsilon}=\underset{\theta\in \overline{\Theta}_{\epsilon}}{\inf} \ \left\|\Phi(\theta)-\Phi(\theta_0)\right\|,\quad  A_{\epsilon}=\{s:\left\|s-\Phi(\theta_0)\right\|\leq \rho_{\epsilon} / 3 \}.\end{align}
Note that $\rho_{\epsilon}>0$ for all $\epsilon >0$ since $\theta\mapsto \Phi(\theta)$ is injective (Assumption~\ref{asmpt:injective_discrete_time}) and continuous (Assumption~\ref{asmpt:statistical_model_discrete_time} and the definition of the random Fourier features), and since $\Theta$ is compact (Assumption~\ref{asmpt:statistical_model_discrete_time}).

We will use the following inequality later on. Denote the distribution of $S_n(X_{1:n}(\theta))$ by $P_{\theta,n}^S$. Observe that, for sufficiently large $n$, 
\begin{align}\label{eqn:inequality_for_set_B_eps}
    P_{\theta_0,n}^S(A_{\epsilon}^c) + \sup_{\theta\in \overline{\Theta}_{\epsilon}} P_{\theta,n}^S(A_{\epsilon}) \leq C n^{-q^{\ast}/2},
\end{align}
because, for sufficiently large $n$, the inequality from~\eqref{eqn:limiting_mean_conv_Phi}
 yields $$P_{\theta_0,n}^S(A_{\epsilon}^c)=P_{\theta_0,n}(\left\|S_n(X_{1:n}(\theta_0))-\Phi(\theta_0)\right\|> \rho_{\epsilon} / 3 )\leq C' (n-m)^{-q^{\ast}/2},$$ for some $C'>0$ and because \begin{align*}&\sup_{\theta\in \overline{\Theta}_{\epsilon}} P_{\theta,n}^S(A_{\epsilon})
\\
&\overset{(1)}{\leq}\sup_{\theta\in \overline{\Theta}_{\epsilon}} P_{\theta,n}(\left\|S_n(X_{1:n}(\theta))-\Phi(\theta)\right\|> \rho_{\epsilon} / 2 )
\\&
\overset{(2)}{\leq}\sup_{\theta\in\Theta} P_{\theta,n}(\left\|S_n(X_{1:n}(\theta))-\Phi(\theta)\right\| > \rho_{\epsilon} / 2 )
\\&
\overset{(3)}{\leq}
C'' n^{-q^{\ast}/2},
\end{align*}
for some $C''>0$ where (1) is by the argument below, (2) is because $\overline{\Theta}_{\epsilon}\subset \Theta$, (3) is by the inequality from~\eqref{eqn:limiting_mean_conv_Phi}. Observe that (1) holds because if $S_n(X_{1:n}(\theta))\in A_{\epsilon}$, then $\left\|S_n(X_{1:n}(\theta))-\Phi(\theta)\right\|>\rho_{\epsilon}/2$, because for every $\theta\in \overline{\Theta}_{\epsilon}$ and $s\in A_{\epsilon}$, we have
\begin{align*}
&    \left\|s-\Phi(\theta)\right\| 
\\& \geq \left|\left\|\Phi(\theta)-\Phi(\theta_0)\right\|-\left\|s-\Phi(\theta_0)\right\|\right|
\\& \geq \left|\rho_{\epsilon} - \rho_{\epsilon}/3\right|
\\& = 2 \rho_{\epsilon}/3,
\end{align*} by adding and subtracting $\Phi(\theta_0)$ and the reverse triangle inequality, the definitions of $\overline{\Theta}_{\epsilon}$ and $A_{\epsilon}$, and subtraction.

\paragraph*{Step 3.2.}

Define  \begin{equation}\label{eqn:def_Pi_M_n}\Pi_n(\overline{\Theta}_{\epsilon}\mid s)=\int_{\overline{\Theta}_{\epsilon}} \pi_n(\theta \mid s) d\theta, \quad M_n(E)=\int_{\Theta} P_{\theta,n}^S(E) \pi(\theta) d\theta,\end{equation} where $P_{\theta,n}^S$ is the distribution of $S_n(X_{1:n}(\theta))$. We aim to show that 
\begin{equation*}\mathbb{E}_{\theta_0,n}\Pi_n(\overline{\Theta}_{\epsilon}\mid S_{n}(X_{1:n}^{\mathrm{obs}}))=o(1).\end{equation*}

Define \begin{equation}\label{eqn:def_Un}U_n = \{\theta \in \Theta:\left\|\theta-\theta_0\right\|\leq n^{-1/2}\},\quad v_n=\int_{U_n} \pi(\theta) d\theta,\end{equation} where $\pi(\theta)$ is the reference density.

Note that $\Theta$ is a Cartesian product of closed bounded intervals (Assumption~\ref{asmpt:statistical_model_discrete_time}) and $\pi(\cdot)$ is continuous and positive (Assumption~\ref{asmpt:ref_dens}). Hence, we have that \begin{equation}\label{eqn:def_v_n}v_n = \int_{U_n} \pi(\theta) d\theta \geq \mu(U_n) \inf_{\theta\in\Theta} \pi(\theta)  \geq  p^{-p/2} n^{-p/2} \inf_{\theta\in\Theta} \pi(\theta)\geq C_v n^{-p/2},\end{equation} for sufficiently large sample size $n\in\mathbb{N}$ and for some fixed $C_v>0$, where $\mu$ denotes the Lebesgue measure. 

Define the mixture distribution \begin{align}\label{eqn:mixture_Qn}Q_n(E)=\frac{1}{v_n}\int_{U_n} P_{\theta,n}^S(E)\pi(\theta)d\theta,\end{align} and observe that
\begin{align*}
    & D_{\mathrm{KL}}(P_{\theta_0,n}^S \| Q_n)
    \\&
    \overset{(1)}{\leq} \frac{1}{v_n} \int_{U_n} D_{\mathrm{KL}}(P_{\theta_0,n}^S \| P_{\theta,n}^S)\pi(\theta)d\theta
    \\&
\overset{(2)}{\leq}    \frac{1}{v_n} \int_{U_n} D_{\mathrm{KL}}(P_{\theta_0,n}\| P_{\theta,n})\pi(\theta)d\theta
    \\&
 \overset{(3)}{\leq} \frac{Cn}{v_n} \int_{U_n} \left\|\theta-\theta_0\right\|^2 \pi(\theta) d\theta 
    \\& 
    \overset{(4)}{\leq} \frac{Cn}{v_n} \int_{U_n} n^{-1} \pi(\theta) d\theta 
    \\&
    \overset{(5)}{=} C,
\end{align*} where (1) convexity of $D_{\mathrm{KL}}$ and since $Q_n$ from~\eqref{eqn:mixture_Qn} is a mixture distribution, (2) by the data processing inequality since $P_{\theta,n}^S=P_{\theta,n}\circ S_n^{-1}$, (3) applying the bound from Assumption~\ref{asmpt:kl_smoothness}, (4) $\left\|\theta-\theta_0\right\|\leq n^{-1/2}$ by the definition of $U_n$ from~\eqref{eqn:def_Un}, (5) by the definition of $v_n$ from~\eqref{eqn:def_Un} and cancelling terms.

Let $\xi>0$ and define \begin{equation}\label{eqn:def_E_n_eps_xi} E_{n,\epsilon,\xi} = A_{\epsilon} \cap \{s:\Pi_n(\overline{\Theta}_{\epsilon}\mid s) > \xi\}.\end{equation}

Observe that
\begin{align*}
&    \xi v_n Q_n(E_{n,\epsilon,\xi})
\\& 
\overset{(1)}{\leq} \xi M_n(E_{n,\epsilon,\xi})
\\& 
\overset{(2)}{\leq} \int_{E_{n,\epsilon,\xi}} \Pi_{n}(\overline{\Theta}_{\epsilon}\mid s) dM_n
\\&
\overset{(3)}{=} \int_{\overline{\Theta}_{\epsilon}} P_{\theta,n}^S(E_{n,\epsilon,\xi}) \pi(\theta) d\theta
\\&
\overset{(4)}{\leq} \sup_{\theta \in \overline{\Theta}_{\epsilon}} P_{\theta,n}^S(A_{\epsilon})
\\&
\overset{(5)}{\leq} C n^{-q^{\ast}/2},
\end{align*}
by (1) the definition of $M_n$ from~\eqref{eqn:def_Pi_M_n}, i.e.,
$$M_n(E)=\int_{\Theta} P_{\theta,n}^S(E) \pi(\theta) d\theta \geq v_n \frac{1}{v_n}\int_{U_n} P_{\theta,n}^S(E)\pi(\theta)d\theta=v_n Q_n(E),$$ since $U_n\subset \Theta$, (2) because $\Pi_n(\overline{\Theta}_{\epsilon}\mid s) > \xi$ by the definition of $E_{n,\epsilon,\xi}$ from~\eqref{eqn:def_E_n_eps_xi}, (3) by the definition of the conditional distributions and Tonelli's theorem to switch the order of integration over $E_{n,\epsilon,\xi}$ and $\overline{\Theta}_{\epsilon}$, (4) $P_{\theta,n}^S(E_{n,\epsilon,\xi})\leq P_{\theta,n}^S(A_{\epsilon})$ since $E_{n,\epsilon,\xi}\subset A_{\epsilon}$ by the definition of $E_{n,\epsilon,\xi}$ from~\eqref{eqn:def_E_n_eps_xi} and upper bounding the integral by the supremum over the domain, factoring out $\sup_{\theta \in \overline{\Theta}_{\epsilon}} P_{\theta,n}^S(A_{\epsilon})$, and noting that $\int_{\overline{\Theta}_{\epsilon}}  \pi(\theta) d\theta\leq 1$, and (5) for some $C>0$ by~\eqref{eqn:inequality_for_set_B_eps}. Hence, we have
$$ Q_n(E_{n,\epsilon,\xi})\leq C \xi^{-1} v_n^{-1} n^{-q^{\ast}/2}\leq   C_v' \xi^{-1}  n^{-(q^{\ast}-p)/2} =o(1),$$by~\eqref{eqn:def_v_n} and since $q^{\ast} > p$. Observe that \begin{equation}\label{eqn:PnS_convergence}P_{\theta_0,n}^S(E_{n,\epsilon,\xi})\leq \frac{C+\log(2)}{\log(1/Q_n(E_{n,\epsilon,\xi}))}=o(1),\end{equation}
by the log sum inequality (Theorem 2.7.1 in \citet{cover_thomas_chapter_2}) and the entropy bound from Theorem 2.6.4 in \citet{cover_thomas_chapter_2}, and since $D_{\mathrm{KL}}(P_{\theta_0,n}^S \| Q_n) \leq C$ and $Q_n(E_{n,\epsilon,\xi})=o(1)$. Note that if $Q_n(E_{n,\epsilon,\xi})=0$, then $P_{\theta_0,n}^S(E_{n,\epsilon,\xi})=0$ because $P_{\theta_0,n}^S \ll Q_n$ since $D_{\mathrm{KL}}(P_{\theta_0,n}^S \| Q_n)<\infty$ as shown above.

Hence, for all $\xi>0$, we have, 
\begin{align*}
&    \mathbb{E}_{\theta_0,n} \Pi_n(\overline{\Theta}_{\epsilon}\mid S_n(X_{1:n}^{\mathrm{obs}}))
\\&
\overset{(1)}{=} \mathbb{E}_{\theta_0,n} [\Pi_n(\overline{\Theta}_{\epsilon}\mid S_n(X_{1:n}^{\mathrm{obs}}))1(\Pi_n(\overline{\Theta}_{\epsilon}\mid S_n(X_{1:n}^{\mathrm{obs}}))\leq \xi)] + \mathbb{E}_{\theta_0,n} [\Pi_n(\overline{\Theta}_{\epsilon}\mid S_n(X_{1:n}^{\mathrm{obs}}))1(\Pi_n(\overline{\Theta}_{\epsilon}\mid S_n(X_{1:n}^{\mathrm{obs}}))> \xi)]
\\&
\overset{(2)}{\leq}  \xi P_{\theta_0,n}(\Pi_n(\overline{\Theta}_{\epsilon}\mid S_n(X_{1:n}^{\mathrm{obs}}))\leq \xi) + P_{\theta_0,n}(\Pi_n(\overline{\Theta}_{\epsilon}\mid S_n(X_{1:n}^{\mathrm{obs}}))> \xi)
\\&
\overset{(3)}{\leq}  \xi + P_{\theta_0,n}(\Pi_n(\overline{\Theta}_{\epsilon}\mid S_n(X_{1:n}^{\mathrm{obs}}))> \xi)
\\&
\overset{(4)}{\leq}  \xi + P_{\theta_0,n}^S(A_{\epsilon}^c) + P_{\theta_0,n}^S(E_{n,\epsilon,\xi}),
\end{align*}
where (1) is by linearity of integration, (2) is from applying the upper bound by $\xi$ on that event for the first term and the upper bound of $1$ for the second term, (3) is since this probability is always less than $1$, (4) follows from the definitions of $\overline{\Theta}_{\epsilon}$ from~\eqref{eqn:def_A}, of $A_{\epsilon}$ from~\eqref{eqn:def_B}, and of $E_{n,\epsilon,\xi}$ from~\eqref{eqn:def_E_n_eps_xi} and then applying a union bound.

Since $\xi>0$ is arbitrary, by the convergence from~\eqref{eqn:inequality_for_set_B_eps} and~\eqref{eqn:PnS_convergence},  we have the desired result that, as $n\xrightarrow[]{}\infty$,
\begin{equation}\label{eqn:convergence_integrated_posterior}
\mathbb{E}_{\theta_0,n}\Pi_n(\overline{\Theta}_{\epsilon}\mid S_{n}(X_{1:n}^{\mathrm{obs}}))=o(1).\end{equation}

\paragraph*{Step 4.}

For each fixed $n\in\mathbb{N}$, we have, as $B\xrightarrow[]{}\infty$, \begin{align*}&
\mathbb{E} \int_{\Theta}
\left|
\hat{\pi}_{n,B}(\theta\mid S_n(X_{1:n}^{\mathrm{obs}}))
-
\pi_n(\theta\mid S_n(X_{1:n}^{\mathrm{obs}}))
\right|d\theta =o(1),\end{align*} by the convergence from Assumption~\ref{asmpt:consistency_conditional_density_estimator}. Hence, we may choose a sequence, $B_n \geq n$, $n\in\mathbb{N}$, that grows sufficiently quickly, such that, for each fixed $n\in\mathbb{N}$, \begin{align*}&
\mathbb{E} \int_{\Theta}
\left|
\hat{\pi}_{n,B_n}(\theta\mid S_n(X_{1:n}^{\mathrm{obs}}))
-
\pi_n(\theta\mid S_n(X_{1:n}^{\mathrm{obs}}))
\right|d\theta  \leq 1/n.\end{align*} With this choice of $B_n$, $n\in\mathbb{N}$, we have \begin{align*} &\mathbb{E}\int_{\overline{\Theta}_{\epsilon}} \hat{\pi}_{n,B_n}(\theta\mid S_n(X_{1:n}^{\mathrm{obs}}))d\theta \\& \leq \mathbb{E}_{\theta_0,n} \Pi_n(\overline{\Theta}_{\epsilon}\mid S_n(X_{1:n}^{\mathrm{obs}})) + \mathbb{E}\int_{\Theta} \left|\hat{\pi}_{n,B_n}(\theta\mid S_{n}(X_{1:n}^{\mathrm{obs}}))-\pi_n(\theta\mid S_{n}(X_{1:n}^{\mathrm{obs}}))\right| d\theta
\\& \leq \mathbb{E}_{\theta_0,n}\Pi_n(\overline{\Theta}_{\epsilon}\mid S_{n}(X_{1:n}^{\mathrm{obs}})) + 1/n
\\& 
=o(1),
\end{align*}
by the convergence result~\eqref{eqn:convergence_integrated_posterior} established in Step 3.

\section{Proof of Corollary~\ref{cor:conv_wass_cond_distr}}

The proof is broken into four steps.

\paragraph*{Step 1.}

Denote the Dirac measure at the true parameter by $\delta_{\theta_0}$. Observe that
$$W_1\left(\hat{\Pi}_{n,B_n},\Pi_n^X\right) \leq W_1\left(\hat{\Pi}_{n,B_n},\delta_{\theta_0}\right) + W_1\left(\Pi_n^X,\delta_{\theta_0}\right),$$ by the triangle inequality for the Wasserstein-1 distance, so it suffices to handle the two terms on the right side.

\paragraph*{Step 2.}
We begin with the first term. Observe that, for any $\epsilon >0$, we have
\begin{align*}W_1\left(\hat{\Pi}_{n,B_n},\delta_{\theta_0}\right) 
& 
\overset{(1)}{=} \int_{\Theta} \left\|\theta-\theta_0\right\| d\hat{\Pi}_{n,B_n}
\\& 
\overset{(2)}{=} \int_{\Theta_{\epsilon}} \left\|\theta-\theta_0\right\| d\hat{\Pi}_{n,B_n}
+ \int_{\overline{\Theta}_{\epsilon}} \left\|\theta-\theta_0\right\|d\hat{\Pi}_{n,B_n}
\\&
\overset{(3)}{\leq} \epsilon \hat{\Pi}_{n,B_n}(\left\|\theta-\theta_0\right\|\leq \epsilon)+ \sup_{\theta_1,\theta_2\in\Theta}\left\|\theta_1-\theta_2\right\| \hat{\Pi}_{n,B_n}(\left\|\theta-\theta_0\right\|>\epsilon)
\\&
\overset{(4)}{\leq} \epsilon + \sup_{\theta_1,\theta_2\in\Theta}\left\|\theta_1-\theta_2\right\| \hat{\Pi}_{n,B_n}(\left\|\theta-\theta_0\right\|>\epsilon),
\end{align*} where (1) is by definition of the Wasserstein-1 distance and Dirac measure at $\theta_0$, i.e., the unique coupling is the product measure $\hat{\Pi}_{n,B_n} \otimes \delta_{\theta_0}$, (2) is by linearity of integration, where $\Theta_{\epsilon}=\{\theta\in\Theta:\left\|\theta-\theta_0\right\|\leq \epsilon\}$, (3) is by upper bounding $\left\|\theta-\theta_0\right\|$ by $\epsilon$ (i.e., by using the definition of $\Theta_{\epsilon}$) and the diameter, in the first and second terms, respectively, (4) is because the measure is upper bounded by $1$.

Taking the expectation of both sides of the inequality yields
\begin{align}\label{eqn:exp_wass_bound}
\mathbb{E}\left[ W_1\left(\hat{\Pi}_{n,B_n},\delta_{\theta_0}\right) \right] \leq    \epsilon + \sup_{\theta_1,\theta_2\in\Theta}\left\|\theta_1-\theta_2\right\| \mathbb{E}\left[\hat{\Pi}_{n,B_n}(\left\|\theta-\theta_0\right\|>\epsilon)\right].
\end{align}  By Theorem~\ref{thm:concentration_of_cond_density_estimator}, for every $\epsilon>0$, we have that 
$$ \mathbb{E}\left[\hat{\Pi}_{n,B_n}(\left\|\theta-\theta_0\right\|>\epsilon)\right]=\mathbb{E}\left[\int_{\overline{\Theta}_{\epsilon}} \hat{\pi}_{n,B_n}(\theta \mid S_n(X_{1:n}^{\mathrm{obs}}))  d\theta \right]=o(1),$$
 for some increasing sequence $B_n$. Therefore, by the inequality from~\eqref{eqn:exp_wass_bound}, we have \begin{equation}\label{eqn:limsup_exp_wass}\limsup_{n\xrightarrow[]{}\infty}\mathbb{E}\left[W_1\left(\hat{\Pi}_{n,B_n},\delta_{\theta_0}\right) \right]\leq    \epsilon,\end{equation} because $\sup_{\theta_1,\theta_2\in\Theta}\left\|\theta_1-\theta_2\right\|$ is finite since $\Theta$ is compact by Assumption~\ref{asmpt:statistical_model_discrete_time}. Since $\mathbb{E}\left[W_1\left(\hat{\Pi}_{n,B_n},\delta_{\theta_0}\right)\right]$ is nonnegative and because~\eqref{eqn:limsup_exp_wass} holds for arbitrary $\epsilon>0$, we have, as $n\xrightarrow[]{}\infty,$  \begin{equation}\label{eqn:lim_exp_wass}\mathbb{E}\left[W_1\left(\hat{\Pi}_{n,B_n},\delta_{\theta_0}\right)\right] = o(1).\end{equation}

This proves the first result. 
 
\paragraph*{Step 3.}

Next, we consider the second term. For any $\epsilon >0$, we have \begin{align*}W_1\left(\Pi_n^X,\delta_{\theta_0}\right) 
& 
\overset{(1)}{=} \int_{\Theta} \left\|\theta-\theta_0\right\| d\Pi_n^X
\\& 
\overset{(2)}{=} \int_{\Theta_{\epsilon}} \left\|\theta-\theta_0\right\| d\Pi_n^X
+ \int_{\overline{\Theta}_{\epsilon}} \left\|\theta-\theta_0\right\|d\Pi_n^X
\\&
\overset{(3)}{\leq} \epsilon \Pi_n^X(\left\|\theta-\theta_0\right\|\leq \epsilon)+ \sup_{\theta_1,\theta_2\in\Theta}\left\|\theta_1-\theta_2\right\| \Pi_n^X(\left\|\theta-\theta_0\right\|>\epsilon)
\\&
\overset{(4)}{\leq} \epsilon+ \sup_{\theta_1,\theta_2\in\Theta}\left\|\theta_1-\theta_2\right\| \Pi_n^X(\left\|\theta-\theta_0\right\|>\epsilon),
\end{align*} where (1) is by definition of the Wasserstein-1 distance and Dirac measure at $\theta_0$, i.e., the unique coupling is the product measure $\Pi_n^X \otimes \delta_{\theta_0}$, (2) is by linearity of integration, where $\Theta_{\epsilon}=\{\theta\in\Theta:\left\|\theta-\theta_0\right\|\leq \epsilon\}$, (3) is by upper bounding by $\epsilon$ and the diameter as in Step 2, and (4) is because the measure is a probability measure so it is upper bounded by $1$.

By the additional assumption in the statement of Corollary~\ref{cor:conv_wass_cond_distr}, we have $$\mathbb{E}\left[\Pi_n^X(\left\|\theta-\theta_0\right\|>\epsilon)\right]=o(1).$$

Hence, we have
$$\mathbb{E}\left[W_1\left(\Pi_n^X,\delta_{\theta_0}\right)\right] = o(1),$$ because $\sup_{\theta_1,\theta_2\in\Theta}\left\|\theta_1-\theta_2\right\|$ is finite since $\Theta$ is compact by Assumption~\ref{asmpt:statistical_model_discrete_time} and because $\epsilon>0$ is arbitrary.

\paragraph*{Step 4.}

Combining the results from Steps 1, 2, and 3, we have the second result that
$$\mathbb{E}\left[W_1\left(\hat{\Pi}_{n,B_n},\Pi_n^X\right)\right] \leq \mathbb{E}\left[W_1\left(\hat{\Pi}_{n,B_n},\delta_{\theta_0}\right)\right] + \mathbb{E}\left[W_1\left(\Pi_n^X,\delta_{\theta_0}\right)\right]=o(1).$$

\end{document}